\documentclass{article}

\usepackage{arxiv}

\usepackage[utf8]{inputenc}
\usepackage[T1]{fontenc}
\usepackage{hyperref}
\usepackage{url}
\usepackage{booktabs}
\usepackage{amsmath}
\usepackage{amsfonts}
\usepackage{amssymb}
\usepackage{nicefrac}
\usepackage{microtype}
\usepackage{cleveref}
\usepackage{graphicx}
\usepackage{natbib}
\usepackage{doi}
\usepackage{multirow}
\usepackage{longtable}
\usepackage{siunitx}

\title{Does Step Law Transfer to \\ Small-Scale Language Models?\\
An Empirical Recalibration Below 59M Parameters}

\newif\ifuniqueAffiliation
\uniqueAffiliationtrue

\author{
  Egor Romanyukov \quad Timofey Novikov \quad Timur Shokarov \\
   \textbf{Elizaveta Zorkina} \quad  \textbf{Anastasia Palienko}  \quad  \textbf{Stepan Dergachev}\\
  \\
  Faculty of Computer Science \\
  HSE University \\
  \texttt{https://github.com/kunikrubika05/step-law-small-scale}
}

\renewcommand{\shorttitle}{Does Step Law Transfer to Small-Scale Language Models?}
\date{}

\hypersetup{
pdftitle={Does Step Law Transfer to Small-Scale Language Models? An Empirical Recalibration Below 59M Parameters},
pdfsubject={cs.LG, cs.CL},
pdfauthor={Elizaveta Zorkina, Timofey Novikov, Egor Romanyukov, Anastasia Palienko, Timur Shokarov},
pdfkeywords={scaling laws, hyperparameter transfer, learning rate, batch size, small language models, Step Law},
}

\begin{document}
\maketitle

\begin{abstract}
Step Law \citep{li2025steplaw} gives power-law formulas for the optimal peak learning rate $\eta^*$ and batch size $B^*$ when pre-training language models:
$\eta^*(N,D) = 1.79\, N^{-0.713} D^{0.307}$ and $B^*(D) = 0.58\, D^{0.571}$.
It was calibrated on models between 59M and 1B parameters; the small-model regime $N < 59M$ was never tested empirically by its authors. This regime matters for single-GPU training, interpretability research, educational experiments, and any setting where a larger model is infeasible on memory or cost grounds.
We test whether Step Law transfers to small language models. We consider three outcomes: H1 -- the original coefficients work directly; H2 -- the power-law form holds but with different coefficients; H3 -- a power law does not describe the optima in this regime. All experiments use a single nanoGPT/TinyStories pipeline with a 2048-token BPE vocabulary, AdamW, and a warmup--cosine schedule. The optimum for each $(N,D)$ cell is extracted from the loss surface $\mathcal{L}(\eta,B)$ via a local quadratic approximation in log--log coordinates over the smoothed training loss.
The final dataset contains 29 unique $(N,D)$ cells and 935 analysis-ready runs. The main refit uses 25 cells (815 runs) in the working range $4 \le D/N \le 600$; two $N=5M$ anchor cells and two heavily data-saturated cells serve only as diagnostics and are excluded from the main regression. On the pooled data we accept H2: the functional form is preserved, but the coefficients differ from the original,
$\hat\eta^*(N,D) = 0.0985\, N^{-0.508} D^{0.238}$ ($R^2=0.834$), $\hat B^*(D) = 3.6\times10^{-4} D^{0.931}$ ($R^2=0.950$).
Step Law's structural claim that $B^*$ is independent of $N$ is reproduced ($p(\log N)=0.87$), but the growth of $B^*$ with $D$ is nearly twice as steep as in the original work. Direct transfer of Step Law systematically overestimates the optimal learning rate: median $\eta_{SL}/\hat\eta \approx 4.0\times$, range $2.4$--$6.6\times$.
\end{abstract}


\section{Introduction}
\label{sec:intro}

Pre-training large language models is among the most expensive procedures in modern machine learning: exhaustive hyperparameter search at full scale is not realistic, and a poor choice of learning rate $\eta$ or batch size $B$ can waste millions of dollars of GPU time. Scaling laws address this problem: empirical relationships that let practitioners predict model behavior from experiments run at a substantially smaller scale. Following Kaplan et al.~\citep{kaplan2020scaling} and Chinchilla~\citep{hoffmann2022chinchilla}, which characterized how loss depends on parameter count $N$ and data budget $D$, the natural next step was to find analogous relationships for the optimization hyperparameters themselves.

Step Law \citep{li2025steplaw}, published in 2025, is currently the most precise result in this direction. Using roughly 3{,}700 training runs, the authors show that $\eta^*$ depends on $N$ and $D$ through a power law, that $B^*$ depends only on $D$, and that the resulting formula predicts the optimum within $0.094\%$ of an exhaustive grid search. Calibration was performed at seven values of $N \in [59M, 1B]$; the region $N < 59M$ was not explored, and the authors explicitly leave generalization beyond the calibrated range to future work \citep[Appendix A.9]{li2025steplaw}.

\paragraph{Why the small-scale regime matters.}
The region $N < 59M$ is not a peripheral corner of the design space. Following TinyStories~\citep{eldan2023tinystories}, it became clear that models with fewer than 10M parameters, trained on a narrow, high-quality corpus, can produce coherent text and display rudimentary reasoning. Small LLMs matter in at least three contexts: edge inference and mobile devices with tight memory and energy budgets; interpretability research, where small models are far easier to analyze mechanistically; and single-GPU training, typical of educational and early-stage projects.

There is also a principled reason to distrust naive extrapolation. Power laws in machine learning routinely break down at the edges of their calibrated range, and reduced model capacity can itself change the geometry of the hyperparameter landscape. Only an experiment can settle whether Step Law applies here.

\paragraph{Contributions.}
We provide a controlled test of Step Law in the region $N < 59M$ on a single nanoGPT/TinyStories/BPE-2048 protocol. For each pair $(N,D)$ we build a loss surface over $(\eta, B)$, extract the empirical optimum, and evaluate three scenarios: direct transfer of the original coefficients, a power-law form with recalibrated coefficients, and rejection of the power-law model altogether. Code, configurations, run logs, and the final tables of optima are published in the project repository.\footnote{\url{https://github.com/kunikrubika05/step-law-small-scale}}

\section{Related Work}
\label{sec:related}

Our work sits at the intersection of four lines of research: scaling laws for language-model loss, empirical laws for optimization hyperparameters, theoretical approaches to hyperparameter transfer across scale, and small language models as an object of study in their own right.

\paragraph{Scaling laws for loss.}
Systematic study of neural language-model scaling begins with Kaplan et al.~\citep{kaplan2020scaling}, who showed that the cross-entropy loss of a transformer depends on $N$, $D$, and compute budget $C$ through a power law across more than seven orders of magnitude, with the practical consequence that large-model behavior can be predicted from small-scale experiments. Hoffmann et al.~\citep{hoffmann2022chinchilla} (Chinchilla) refined this picture: at a fixed compute budget, the optimal ratio of tokens to parameters is roughly 20:1. Both works describe how loss depends on scale and data but are silent on what the optimization hyperparameters $\eta$ and $B$ should be for a given $(N,D)$. These scaling laws are themselves only valid inside their calibrated range: extrapolating far outside it, e.g.\ to very small or very large $N$, is known to produce deviations from the power-law regime~\citep{kaplan2020scaling}.

\paragraph{Optimizer, warmup, and the learning-rate schedule.}
The choice of optimizer and learning-rate schedule directly determines which $\eta$ turns out to be optimal; without fixing them, comparing Step Law's predictions against empirical optima is not meaningful. Kingma and Ba~\citep{kingma2015adam} introduced Adam, which adapts the step size per parameter; Loshchilov and Hutter~\citep{loshchilov2019decoupled} showed that $L_2$ regularization behaves differently from weight decay under Adam and proposed AdamW, now standard for pre-training transformers. Step Law's own schedule is likewise fixed: linear warmup over 2{,}000 steps followed by cosine decay to $\eta_{\min}=10^{-5}$. The length of warmup and the value of $\eta_{\min}$ are known to affect the shape of the loss landscape~\citep{liu2020variance}, so we retain the same class of schedule -- warmup followed by cosine decay -- in our experiments.

\paragraph{Batch size, gradient noise, and the critical batch size.}
The critical batch size $B^*$ is a central concept for this work. Keskar et al.~\citep{keskar2017largebatch} showed that large-batch training tends to converge to ``sharp'' minima with worse generalization, whereas small-batch training finds ``flat'' minima that generalize better -- establishing batch size as a genuine hyperparameter rather than a mere parallelization knob. McCandlish et al.~\citep{mccandlish2018empirical} gave a theoretical grounding for $B^*$ through the gradient noise scale $\mathcal{B}$, the ratio between the variance of the gradient and the norm of the gradient itself: for $B \ll \mathcal{B}$, increasing batch size accelerates training nearly linearly, while for $B \gg \mathcal{B}$ the gain becomes negligible. This methodology was used, for example, to choose the batch size for GPT-3. The strong assumptions behind gradient noise scale (quadratic loss geometry, isotropic noise) hold poorly in practice. Merrill et al.~\citep{merrill2025criticalbatch} recently proposed replacing gradient noise scale with a short series of probe runs at different batch sizes, without assuming anything about loss geometry -- calling into question the standard McCandlish-style reasoning that Step Law itself relies on, and making the behavior of $B^*$ at small scale even less certain a priori.

\paragraph{Empirical laws for optimization hyperparameters.}
Transferring the logic of scaling laws from loss to the hyperparameters themselves is a comparatively recent effort. DeepSeek-AI~\citep{deepseek2024llm} built power laws for $\eta^*$ and $B^*$ as functions of compute budget $C=ND$ and validated their predictive power at 7B and 67B scale. This was an important step, but folding the dependence on $N$ and $D$ into a single quantity $C$ hides possible asymmetries -- for instance, a small model trained on an unusually large corpus. Step Law~\citep{li2025steplaw} makes this dependence explicit. Using roughly 3{,}700 runs at seven values of $N$ from 59M to 1B, the authors show that $B^*$ depends only on $D$, not on $N$, and that the resulting formula predicts hyperparameters within a $0.094\%$ relative deviation from the global optimum. Outside $N \in [59M, 1B]$ the law was not tested empirically; the authors explicitly defer generalization beyond the validated range to future work \citep[Appendix A.9]{li2025steplaw}.

\paragraph{Hyperparameter transfer via parameterization.}
An alternative, theory-driven approach is maximal update parameterization ($\mu$P), proposed by Yang et al.~\citep{yang2022tensorprograms}. Under $\mu$P, the optimal learning rate is approximately invariant to model width: hyperparameters found on a $\sim$40M proxy model transfer to a 6.7B GPT-3-scale model at a cost of roughly 7\% additional compute. The approach is elegant but requires architectural modifications (special initialization scales, per-parameter learning rates) and does not model the dependence of $\eta$ on data volume. Step Law, by contrast, is a purely empirical formula on a standard parameterization, which makes it more attractive to practitioners who do not want to rewrite their architecture. $\mu$P and Step Law are complementary rather than competing: $\mu$P describes how $\eta^*$ behaves with width $N$ at fixed $D$, while Step Law additionally supplies the dependence on $D$. Our experiments in the $N < 59M$ regime, on a standard (non-$\mu$P) parameterization, let us compare empirical observations against the predictions of both approaches.

\paragraph{Small language models.}
The region $N < 59M$ is not peripheral, practically or scientifically. Eldan and Li~\citep{eldan2023tinystories} showed that language models with 1--30M parameters, trained on a specially constructed corpus of synthetic short stories, produce coherent text with causal structure. This line of work on ``small language models'' suggests that small models are not merely scaled-down copies of large ones, but behave as qualitatively different systems with a potentially different hyperparameter landscape geometry. Mechanistic interpretability~\citep{elhage2021mathematical} uses small transformers as its primary object of study: phenomena such as circuits, superposition, and induction heads were first described at the 1--10M parameter scale. Understanding the optimal training regime at this scale matters for the reproducibility of such results. In edge inference and single-GPU training, small models are the only realistic option, and reliable guidance on choosing $\eta$ and $B$ is needed here as well.

\paragraph{Research gap.}
Table~\ref{tab:related} summarizes the key related work along two axes: the range of $N$ covered, and whether the work models loss or hyperparameters. The scaling laws of Kaplan and Chinchilla cover a wide range of $N$ but describe only loss. $\mu$P covers a wide range of $N$ for $\eta$, but on a modified parameterization and without a dependence on $D$. Step Law~\citep{li2025steplaw} is the first work to give a joint dependence $\eta^*(N,D)$ and $B^*(D)$ on a standard parameterization, but only for $N \ge 59M$. The region $N < 59M$, on a standard parameterization, with a joint dependence of hyperparameters on both $N$ and $D$, remains unexplored -- this is the gap we address.

\begin{table}[h]
\centering
\caption{Comparison with existing work.}
\label{tab:related}
\begin{tabular}{lccc}
\toprule
Work & Range of $N$ & Target & Dependence on $D$ \\
\midrule
Kaplan et al.~\citep{kaplan2020scaling} & $10^{7}$--$10^{10}$ & Loss & Yes \\
Chinchilla~\citep{hoffmann2022chinchilla} & $10^{8}$--$10^{11}$ & Loss & Yes \\
McCandlish et al.~\citep{mccandlish2018empirical} & $10^{7}$--$10^{9}$ & $B^*$ & Indirect \\
$\mu$P~\citep{yang2022tensorprograms} & $10^{7}$--$10^{10}$ & $\eta^*$ & No \\
DeepSeek~\citep{deepseek2024llm} & $10^{9}$--$10^{11}$ & $\eta^*, B^*$ & Via $C$ \\
Step Law~\citep{li2025steplaw} & $6\times10^{7}$--$10^{9}$ & $\eta^*, B^*$ & Explicit \\
\textbf{This work} & $2.5\times10^{5}$--$2\times10^{6}$ & $\eta^*, B^*$ & Explicit \\
\bottomrule
\end{tabular}
\end{table}

\section{Problem Setup and Hypotheses}
\label{sec:setup}

\subsection{Notation}
\label{sec:notation}

Throughout, $N$ denotes the number of model parameters (excluding the embedding layer), $D$ the training data budget in tokens, $\eta$ the peak learning rate under a warmup$\to$cosine schedule, and $B$ the batch size in tokens. We write $\mathcal{A}$ for the architecture space, $\mathcal{D}$ for the data distribution, and $\mathcal{L}(\mathcal{A}, \mathcal{D}, N, D, \eta, B)$ for the loss. The main optimality metric is the smoothed training loss $\mathcal{L}_{\text{smooth}}$, the training loss averaged over the last steps of training; its precise definition and justification as a proxy for the optimum are given in Section~\ref{sec:metric}.

A \emph{run} is a single training job at fixed $N, D, \eta, B$, and seed. A \emph{sweep} is a set of runs over a predefined hyperparameter grid -- in our setting, a grid search over $(\eta, B)$ for a chosen pair $(N,D)$. Implementation details (run configuration, logging, storage) are relegated to the project repository; in the main text a run is treated as an abstract experimental object $(N, D, \eta, B, \text{seed}) \mapsto \mathcal{L}_{\text{smooth}}$.

\subsection{Formal Statement}
\label{sec:formal}

For fixed $\mathcal{A}, \mathcal{D}, N, D$, the optimal hyperparameters are defined as
\begin{equation}
\eta^*, B^* = \arg\min_{\eta, B} \mathcal{L}_{\mathcal{A}, \mathcal{D}, N, D}(\eta, B). \label{eq:argmin}
\end{equation}
Empirically, in log--log coordinates, the optima are well described by the power-law family
\begin{equation}
\eta^*(N,D) = c \cdot N^\alpha \cdot D^\beta, \qquad B^*(D) = d \cdot D^\gamma, \label{eq:family}
\end{equation}
where $c, \alpha, \beta, d, \gamma$ are fitted constants, and $B^*$ depends only on $D$, not on $N$. The specific instantiation of family~\eqref{eq:family} calibrated on large models (Step Law) is given in Section~\ref{sec:steplaw}.

We study problem~\eqref{eq:argmin} in the small-model regime
\begin{equation}
N \in (0, N_{\min}), \qquad N_{\min} = 59M \text{ parameters}, \label{eq:regime}
\end{equation}
at several values of $D$. For each pair $(N,D)$ on a grid within~\eqref{eq:regime} we find the empirical optimum
\begin{equation}
\hat\eta(N,D), \hat B(N,D) = \arg\min_{\eta,B} \mathcal{L}_{\text{smooth}}(\eta,B) \label{eq:empirical}
\end{equation}
by grid search.

\paragraph{Research question.} The power-law family of Step Law is calibrated on $N \ge N_{\min}$, where it describes the optima with a relative deviation of $\approx 0.094\%$. Our central question is whether this description transfers to regime~\eqref{eq:regime}: (a) do the original coefficients hold; (b) does at least the power-law form hold; and (c) what is the cost of a mismatch, in terms of excess loss. The region $N < N_{\min}$ was never tested empirically by the authors \citep[Appendix A.9]{li2025steplaw}. Power laws in machine learning have a limited domain of validity, and reduced model capacity at small $N$ may qualitatively change the geometry of the hyperparameter landscape, so direct extrapolation of formula~\eqref{eq:family} is not guaranteed \emph{a priori}.

\subsection{Hypotheses}
\label{sec:hypotheses}

We formulate three pairwise mutually exclusive and exhaustive hypotheses. Let $(\eta_{SL}, B_{SL})$ denote Step Law's predictions with the original coefficients (Section~\ref{sec:steplaw}).

\paragraph{H1 -- The original Step Law coefficients hold for small models.}
Formally, for all $N < N_{\min}$ and the $D$ values under study,
\begin{equation}
\hat\eta(N,D) \approx 1.79 \cdot N^{-0.713} \cdot D^{0.307}, \qquad \hat B(D) \approx 0.58 \cdot D^{0.571}, \label{eq:h1}
\end{equation}
with quality assessed via the metric $\delta(N,D)$ from Section~\ref{sec:metric}: we compare the smoothed training loss at the point predicted by Step Law against the best smoothed training loss from grid search. The reference figure of $0.094\%$ is taken from the original paper as the excess loss relative to the global optimum \citep[Section 3.4.3]{li2025steplaw}. If true, Step Law would be genuinely scale-invariant, and the original formulas could be applied directly when pre-training small LLMs, without recalibration.

\paragraph{H2 -- The power-law form holds, but with different coefficients.}
Formally, there exist $c', \alpha', \beta', d', \gamma'$ such that
\begin{equation}
\hat\eta(N,D) \approx c' \cdot N^{\alpha'} \cdot D^{\beta'}, \qquad \hat B(D) \approx d' \cdot D^{\gamma'}, \label{eq:h2}
\end{equation}
with at least one coefficient differing from the values in Eq.~\eqref{eq:steplaw}. If true, the power-law structure of the dependence of $\eta^*$ and $B^*$ on $N$ and $D$ would be a fundamental property of language-model optimization, but the specific coefficients would depend on scale, requiring a separate calibration for small models. The fitted coefficients, together with code, configurations, and experiment logs, are released in an open benchmark (project repository).

\paragraph{H3 -- No power law holds at small scale.}
Formally, there is no power law $\hat\eta(N,D) = c \cdot N^\alpha \cdot D^\beta$ that fits the empirical optima with acceptable accuracy over regime~\eqref{eq:regime}. If true, the hyperparameter landscape at small $N$ would be qualitatively different from the large-model regime, pointing to the need for alternative functional forms (e.g.\ piecewise power laws with a transitional regime, or exponential forms) and implying that existing scaling laws cannot be mechanically transferred to small models.

These three hypotheses are pairwise mutually exclusive and jointly exhaustive with respect to the possible outcomes of testing formula~\eqref{eq:steplaw} in regime~\eqref{eq:regime}.

\section{Background: Step Law}
\label{sec:steplaw}

Step Law~\citep{li2025steplaw} gives a specific instantiation of the power-law family~\eqref{eq:family}. Using the notation of \citet[Appendix A.1]{li2025steplaw} and the optimality definition~\eqref{eq:argmin} (their Definition 1), the authors fit coefficients by least squares and obtain
\begin{equation}
\eta^*(N,D) = 1.79 \cdot N^{-0.713} \cdot D^{0.307}, \qquad B^*(D) = 0.58 \cdot D^{0.571}. \label{eq:steplaw}
\end{equation}
Their regression analysis \citep[Appendix A.5]{li2025steplaw} confirms that $B^*$ does not depend on $N$. The coefficients were fit on 1{,}912 runs at seven values of $N$ from 59M to 1B and five values of $D$ (2B, 4B, 8B, 20B, and 100B tokens) \citep[Section 3.4]{li2025steplaw}; within this range, formula~\eqref{eq:steplaw} predicts the optimum with a relative deviation of $\approx 0.094\%$. These $(\eta_{SL}, B_{SL})$ serve as the reference point for testing H1.

\paragraph{Learning-rate schedule.}
The authors use a standard scheduler: linear warmup over the first 2{,}000 steps, followed by cosine decay to a fixed $\eta_{\min} = 10^{-5}$ \citep[Section 3.3.2]{li2025steplaw}. Under this schedule, $\eta$ in Eq.~\eqref{eq:steplaw} is specifically the peak learning rate. Since the schedule is standardized across all their experiments, Step Law predicts only the peak, not the full trajectory. Because the shape of the landscape is sensitive to the schedule, we reproduce the same class of scheduler in our own experiments.

\paragraph{Smoothed training loss as a proxy for the optimum.}
Step Law uses smoothed training loss as its optimality metric. The authors show that it is an unbiased proxy for validation loss: both are minimized at the same values of $\eta$ and $B$ \citep[Section 3.3.3]{li2025steplaw}. This justification carries over to our setting and validates both the H1 comparison and the estimation of power-law coefficients from smoothed loss (Section~\ref{sec:metric}).

\paragraph{The structure of $B^*$ and gradient noise.}
The claim that $B^*$ depends only on $D$ is consistent with the notion of gradient noise scale~\citep{mccandlish2018empirical}: the critical batch size is set by the ratio of gradient noise to gradient norm, which changes primarily over the course of training (i.e.\ with $D$) rather than with model width. We test this structural claim separately (metric M4 in Section~\ref{sec:metric}), since it is conceptually more important than the exact value of the exponent.

\section{Experiments}
\label{sec:experiments}

This section describes the final experimental protocol and the results on the pooled data. We use all successfully completed runs, normalize the $(N,D)$ cells, and estimate the power-law coefficients on the main refit grid. The training loop, configurations, logs, aggregation scripts, and final tables are released in the project repository.

\subsection{Protocol and Final Grid}
\label{sec:protocol}

\paragraph{Models and data.}
We work in the small-model regime $N < 59M$, on a nanoGPT-style architecture in the standard, non-$\mu$P, parameterization. The main scale range is $N \in \{255K, 519K, 997K, 2.03M\}$. For diagnostics, we additionally ran two anchor cells at $N=5M$, which are excluded from the main refit so as not to mix this diagnostic, larger scale with the dense main grid. The data is roneneldan/TinyStories~\citep{eldan2023tinystories}; tokenization uses a single in-domain BPE tokenizer with a 2{,}048-token vocabulary.

\paragraph{Final $(N,D)$ grid.}
After merging iterations 1 and 2, we obtain 29 unique $(N,D)$ cells and 935 analysis-ready runs. Two runs without a \texttt{finish}/step status were excluded; there are no duplicate hyperparameter points $(N,D,\eta,B,\text{seed})$. The main regression uses 25 cells (815 runs) in the working range $4 \le D/N \le 600$: each of the four main scales has at least four values of $D$, which makes the estimated exponents in $N$ and $D$ more stable than in the first iteration. The cells $N{=}255K/D{=}537M$ and $N{=}519K/D{=}537M$ are kept as anchor/diagnostic cells because $D/N > 600$, and the two $N{=}5M$ cells are kept as a separate B1-anchor set outside the main refit.

\begin{table}[h]
\centering
\caption{Summary of the final grid. The main refit excludes $N{=}5M$ and cells with $D/N > 600$; all cells are retained for diagnostics.}
\label{tab:grid}
\begin{tabular}{lcc}
\toprule
Data group & $(N,D)$ cells & Usable runs \\
\midrule
All unique cells & 29 & 935 \\
Iteration 1 & 12 & 405 \\
Iteration 2 (excl. $N{=}5M$) & 15 & 480 \\
$N{=}5M$ anchor & 2 & 50 \\
\textbf{Main refit} & \textbf{25} & \textbf{815} \\
\bottomrule
\end{tabular}
\end{table}

\paragraph{$(\eta,B)$ grid and training.}
For each cell $(N,D)$ we run a grid search over the peak learning rate $\eta$ and batch size $B$ (in tokens). The center of the $\eta$ window was set by the Step Law prediction, but whenever the minimum fell on the lower boundary the window was extended downward; already in the first iteration it was clear that the original formula systematically overestimates $\eta^*$ at small scale. Batch size was scaled together with $D$ in steps of $\times 2$. All runs use AdamW ($\beta_1=0.9$, $\beta_2=0.95$, weight decay $0.1$, gradient clipping $1.0$), bf16, and a single warmup--cosine scheduler with warmup equal to 5\% of the budget and $\eta_{\min}=10^{-5}$. A single seed, 1337, was used throughout; the statistical intervals reported below therefore reflect variation across cells rather than initialization noise.

\subsection{Design and Departures from the Step Law Setup}
\label{sec:design}

Testing Step Law at our scale is not a literal replication of the original study. We preserve the object of comparison -- the location of the optimum $(\hat\eta,\hat B)$ in hyperparameter coordinates -- but adapt the experiment to models tens to hundreds of times smaller than the authors' range.

\begin{itemize}
\item \textbf{Scale.} The authors calibrate their law at $N \ge 59M$, whereas our main range is 0.25--2.03M parameters.
\item \textbf{Tokenizer.} Instead of the $\sim$50K-token GPT-2 BPE vocabulary, we use a 2{,}048-token BPE tokenizer trained on TinyStories. This avoids a regime where the embedding/output projection would dominate a tiny transformer backbone.
\item \textbf{Warmup.} Step Law fixes warmup at 2{,}000 steps. In our grid, the number of steps varies as $D/B$, so a fixed 2{,}000-step warmup would correspond to incomparable fractions of training across cells. We keep the warmup--cosine \emph{class} of schedule but set warmup to 5\% of the training budget.
\item \textbf{Single seed.} Our final experiments have no replicates over initialization. This is our main limitation: we can assess stability across cells and the shape of the loss surface, but not a genuine seed-level confidence interval.
\end{itemize}

\subsection{Tokenizer Choice and Its Consequences}
\label{sec:tokenizer}

At $N$ on the order of 0.25--2.0M parameters, the GPT-2 tokenizer introduces a substantial confounder: a 50K-token vocabulary makes the embedding and output layers disproportionately large relative to the transformer blocks. Under such a setup, the experiment would end up measuring not only the scaling of the backbone model's optimal hyperparameters, but also the cost of an oversized vocabulary projection. We therefore fix, for the main grid, a single in-domain BPE tokenizer with a 2{,}048-token vocabulary trained on TinyStories.

Absolute cross-entropy values between BPE-2048 and GPT-2 BPE are not directly comparable, because their tokens carry different amounts of information. We therefore do not compare absolute loss values to the numbers reported by Step Law, but only compare the locations of the optima $(\hat\eta,\hat B)$ and the power-law exponents at a fixed tokenizer, within our own grid. Early GPT-2-tokenizer smoke runs were used only as an infrastructure check; the final H1/H2/H3 conclusions are based entirely on the unified BPE-2048 setup.

\subsection{Metric and Optimum Extraction}
\label{sec:metric}

The main metric is smoothed training loss, following Step Law~\citep{li2025steplaw}. For each completed run, we take the smoothed training loss at the end of training:
\begin{equation}
\mathcal{L}_{\text{smooth}}(\eta,B) = \frac{1}{K}\sum_{t=T-K+1}^{T}\mathcal{L}_t(\eta,B), \qquad K=128. \label{eq:smooth}
\end{equation}
For each cell $(N,D)$ we build the surface $\mathcal{L}(\eta,B)$ over all completed runs. The optimum is extracted with a \emph{robust 2D-quadratic} method: within the window $\mathcal{L} \le \mathcal{L}_{\min}+0.15$, we fit a local quadratic approximation
\begin{equation}
\mathcal{L} \approx q(\log\eta, \log B). \label{eq:quad}
\end{equation}
If the Hessian is positive definite and the vertex lies inside the explored box, the vertex is taken as $(\hat\eta,\hat B)$; otherwise we fall back to the observed grid minimum. On the 12 cells of the first iteration, this method exactly reproduces the previously recorded optima, so results from different parts of the grid are directly comparable.

To test the power-law form, we then fit OLS regressions in log--log coordinates:
\begin{equation}
\log\hat\eta = \log c - \alpha \log N + \beta \log D, \qquad \log\hat B = \log d + \gamma \log D. \label{eq:ols}
\end{equation}
For the structural claim about batch size, we additionally test the extended model $\log\hat B = \log d + a\log N + \gamma\log D$ and run a hierarchical $F$-test on the coefficient of $\log N$.

\subsection{Final Results}
\label{sec:results}

\subsubsection{Direct Transfer of Step Law: Systematically Overestimated Learning Rate}
\label{sec:overshoot}

The main effect is robust to merging the two iterations: Step Law's original formula predicts a peak learning rate that is far too large. Across the 25 cells of the main refit, the median ratio $\eta_{SL}/\hat\eta$ is about $4.0\times$, with a range of $2.4$--$6.6\times$. The overshoot decreases with growing $N$, consistent with our $N$-dependence for $\eta^*$ being weaker than the original.

\begin{table}[h]
\centering
\caption{Overestimation of the optimal learning rate by the Step Law formula, per main-refit scale.}
\label{tab:overshoot}
\begin{tabular}{lcc}
\toprule
Scale $N$ & Geometric mean $\eta_{SL}/\hat\eta$ & Per-cell range \\
\midrule
0.25M & $4.97\times$ & $3.9$--$6.6\times$ \\
0.52M & $3.83\times$ & $2.8$--$4.2\times$ \\
1.0M  & $3.69\times$ & $2.4$--$6.2\times$ \\
2.0M  & $3.28\times$ & $2.8$--$4.1\times$ \\
\bottomrule
\end{tabular}
\end{table}

\begin{figure}[h]
\centering
\includegraphics[width=\textwidth]{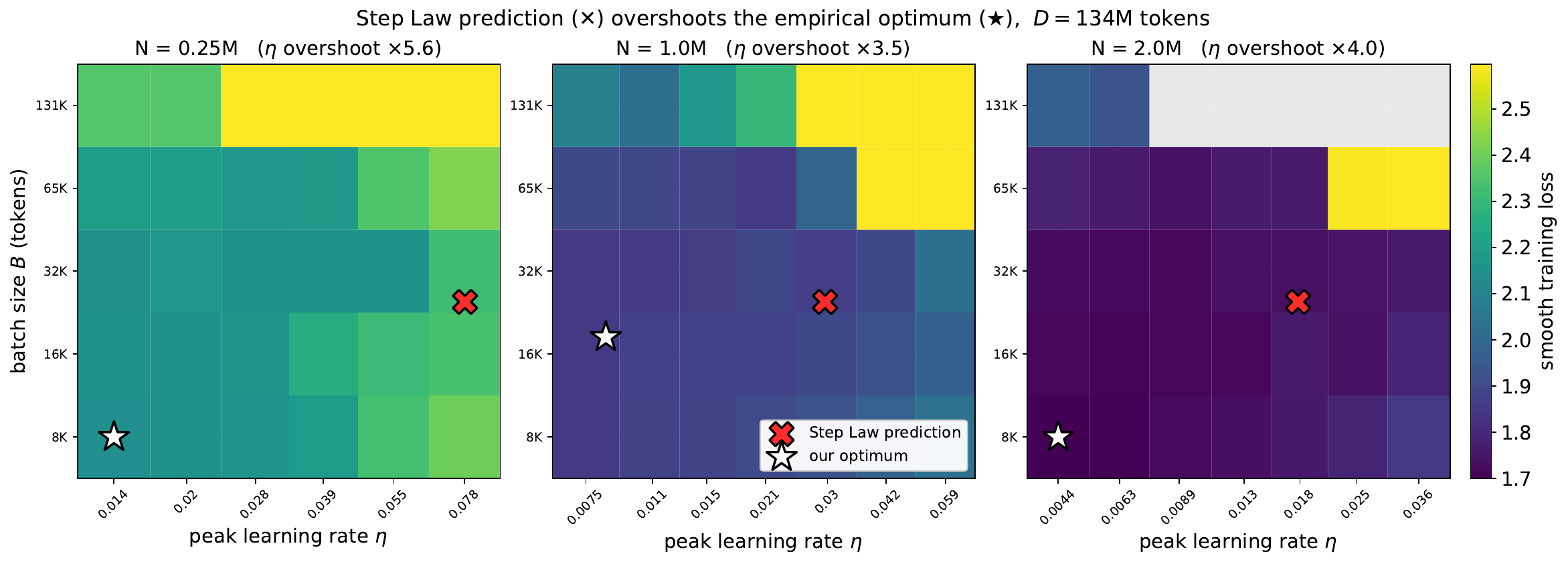}
\caption{Example hyperparameter landscapes at $D=134M$ tokens. The red cross marks the Step Law prediction, the star marks the empirical robust optimum. At every scale shown, the Step Law prediction is shifted toward an overestimated learning rate.}
\label{fig:landscape}
\end{figure}

\subsubsection{Recalibrated Coefficients: H2 Instead of H1}
\label{sec:refit}

On the 25 main cells, the final refit gives
\begin{align}
\hat\eta^*(N,D) &= 0.0985\, N^{-0.508} D^{0.238}, & R^2 &= 0.834,\ \text{adj-}R^2 = 0.819, \label{eq:eta-refit}\\
\hat B^*(D) &= 3.6\times10^{-4}\, D^{0.931}, & R^2 &= 0.950. \label{eq:b-refit}
\end{align}
Both coefficients in the $\eta^*$ model are significant. The $N$-exponent is stable across subsamples and substantially shallower in magnitude than the original: $0.508$ versus $0.713$. The $D$-exponent for $\eta^*$ is $0.238$, statistically compatible with the original $0.307$ at the margin of significance. The strongest discrepancy is in batch size: $\gamma = 0.931$ versus $0.571$ for Step Law.

\begin{table}[h]
\centering
\caption{Final power-law exponents compared with Step Law's coefficients. The bootstrap CI is obtained by resampling the 25 cells.}
\label{tab:coefficients}
\begin{tabular}{lcccc}
\toprule
Exponent & Our estimate & 95\% bootstrap CI & Step Law & $p$ (difference) \\
\midrule
$\alpha$ in $\eta^* \propto N^{-\alpha}$ & 0.508 & $[0.410, 0.606]$ & 0.713 & $5\times10^{-5}$ \\
$\beta$ in $\eta^* \propto D^{\beta}$ & 0.238 & $[0.159, 0.300]$ & 0.307 & $0.051$ \\
$\gamma$ in $B^* \propto D^{\gamma}$ & 0.931 & $[0.860, 0.994]$ & 0.571 & $2\times10^{-24}$ \\
\bottomrule
\end{tabular}
\end{table}

\begin{figure}[h]
\centering
\includegraphics[width=0.85\textwidth]{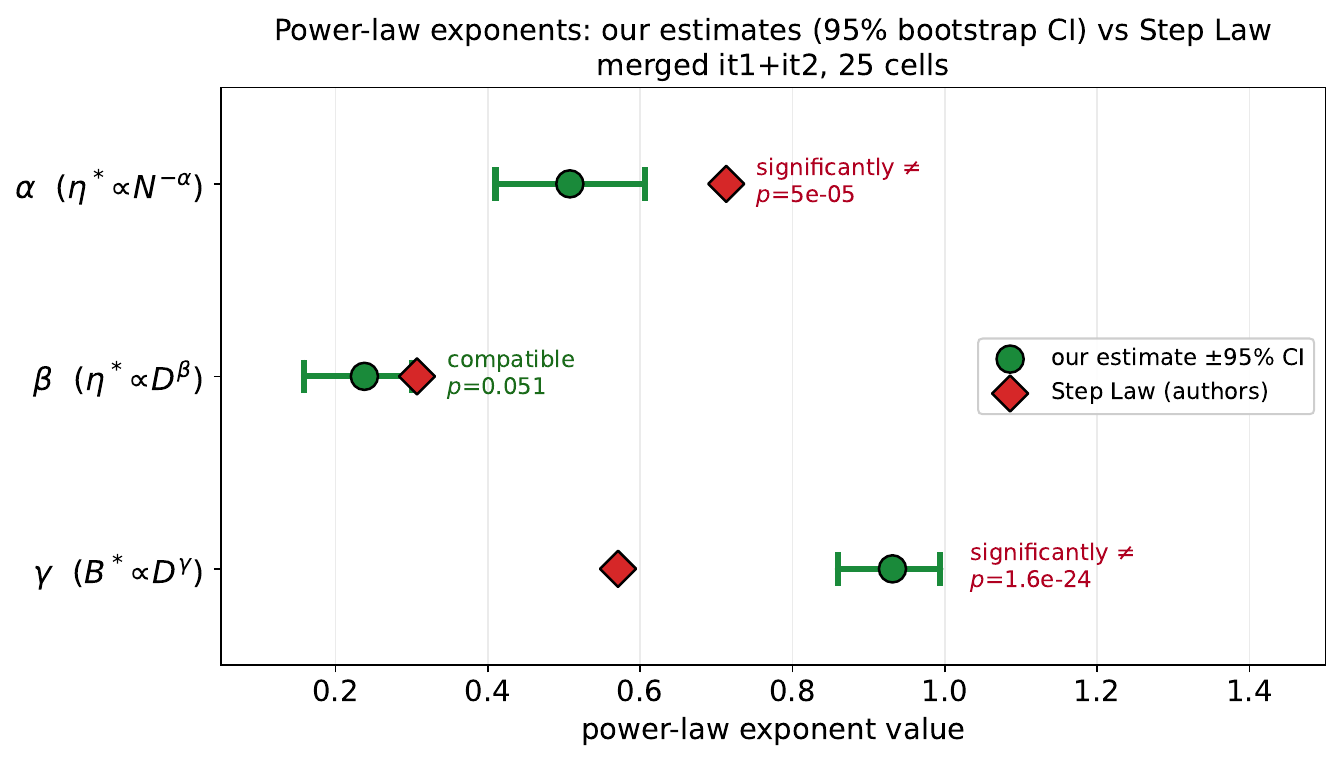}
\caption{Power-law exponents: our estimates with 95\% bootstrap CIs against Step Law's coefficients. $\alpha$ is significantly shallower, $\beta$ is close to the original, $\gamma$ for batch size is significantly steeper.}
\label{fig:coefficients}
\end{figure}

\subsubsection{Batch Size Depends on $D$, Not on $N$}
\label{sec:bstar}

We separately test Step Law's structural claim: the optimal batch size depends on the data budget but not on the parameter count. A model that adds $\log N$ for batch size gives
\begin{equation}
\hat B^* = 4.1\times10^{-4}\, N^{-0.013} D^{0.933}, \qquad R^2=0.950, \label{eq:b-extended}
\end{equation}
where the coefficient on $\log N$ is statistically indistinguishable from zero: $p(\log N)=0.87$. A hierarchical $F$-test likewise finds no improvement from adding $\log N$ ($F=0.03$, $p=0.87$). The structural part of Step Law therefore reproduces, even though the exponent $\gamma$ on $D$ differs substantially.

\begin{figure}[h]
\centering
\includegraphics[width=0.85\textwidth]{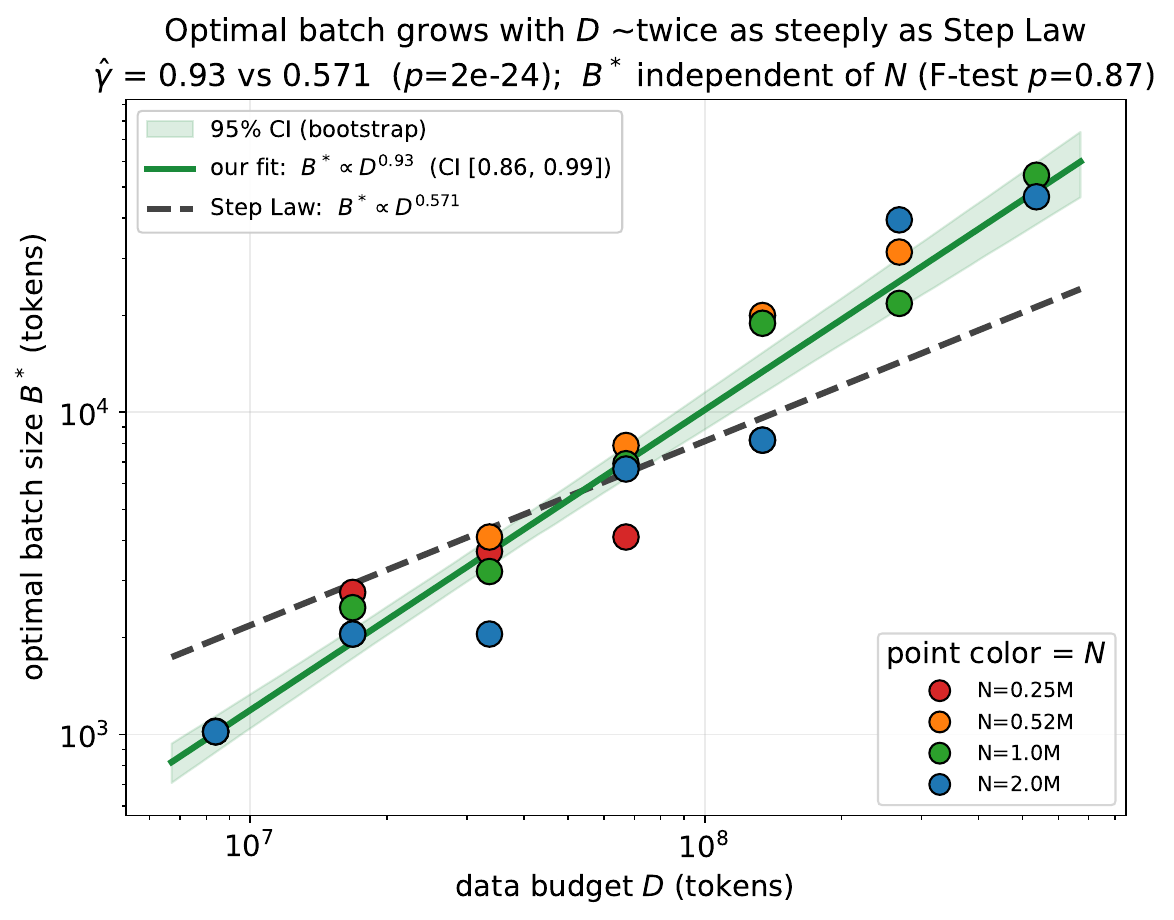}
\caption{Optimal batch size grows with the data budget $D$ as a power law. Point color denotes scale $N$; the absence of systematic separation by color illustrates the independence of $B^*$ from $N$.}
\label{fig:bstar}
\end{figure}

\subsubsection{Stability Across Iterations}
\label{sec:stability}

The final merged refit does not overturn the conclusions of the second iteration. On the contrary, the first and second iterations give a consistent picture: $\alpha$ stays around $0.51$--$0.55$, $\gamma$ around $0.93$--$1.02$, and the independence of $B^*$ from $N$ is preserved. The most sensitive quantity is the $D$-exponent $\beta$ for the learning rate, which ranges from $0.20$ to $0.35$ across subsamples -- expected given a single seed and noisy optimum locations.

\begin{table}[h]
\centering
\caption{Stability of the coefficients across subsamples. For $\eta^*$, the form $\eta^* \propto N^{-\alpha}D^{\beta}$ is shown.}
\label{tab:stability}
\begin{tabular}{lcccccc}
\toprule
Subsample & $n$ & $\alpha$ (on $N$) & $\beta$ (on $D$) & $R^2(\eta)$ & $\gamma$ & $R^2(B)$ \\
\midrule
Iteration 1 only & 12 & 0.548 & 0.350 & 0.954 & 1.018 & 0.938 \\
Iteration 2 only (excl.\ $N{=}5M$) & 15 & 0.529 & 0.198 & 0.791 & 0.935 & 0.967 \\
Merged main & 25 & 0.508 & 0.238 & 0.834 & 0.931 & 0.950 \\
All non-$N{=}5M$ & 27 & 0.550 & 0.274 & 0.871 & -- & -- \\
Step Law & -- & 0.713 & 0.307 & -- & 0.571 & -- \\
\bottomrule
\end{tabular}
\end{table}

\subsection{Verdict on H1/H2/H3}
\label{sec:verdict}

\begin{table}[h]
\centering
\small
\caption{Final verdict on the three hypotheses, on the pooled data.}
\label{tab:verdict}
\begin{tabular}{p{0.08\textwidth}p{0.28\textwidth}p{0.42\textwidth}p{0.12\textwidth}}
\toprule
Hyp.\ & What is tested & Final result & Verdict \\
\midrule
H1 & Original Step Law coefficients hold at $N<59M$ & $\eta_{SL}$ overestimates the optimum by a median of $\approx4.0\times$; $\alpha$ and $\gamma$ differ significantly from the original; iteration 1 already showed a median loss gap of $3.6\%$ & Rejected \\
H2 & Power-law form holds, coefficients differ & $\eta^*$ is well described by a power law in $N,D$ ($R^2=0.834$); $B^*$ by a power law in $D$ ($R^2=0.950$); $\log N$ for $B^*$ is not significant ($p=0.87$) & Accepted \\
H3 & Power law does not describe small scale & High $R^2$, consistent iterations, and a reproducible structure of $B^*(D)$ give no grounds to reject the power-law form & Not accepted \\
\bottomrule
\end{tabular}
\end{table}

The substantive conclusion is that small scale does not break the power-law structure itself, but it does change the calibration. For training small LLMs in practice, Step Law should not be used verbatim: a safe heuristic based on our data is to use the recalibrated formula~\eqref{eq:eta-refit}, or, roughly, to take about one quarter of the learning rate that Step Law prescribes.

\subsection{Limitations}
\label{sec:limitations}

The main limitation of this final version is the single seed. The robust 2D-quadratic fit uses the shape of the surface and is therefore less sensitive to a single lucky point than a plain observed minimum, but it does not create genuine replicates over initialization. The bootstrap CIs in Table~\ref{tab:coefficients} should therefore be read as variation across cells, not as a full confidence interval over seeds. The $D$-exponent $\beta$ for the learning rate deserves particular caution: it is statistically compatible with the original value, but sits at the margin of significance and remains the most sensitive of the reported exponents.

The second limitation is range. The main refit covers 0.25--2.03M parameters, well below Step Law's lower bound of 59M. The two $N=5M$ cells are used only as anchors. Conclusions should therefore be restricted to the studied range $N < 2.03M$; the $N=5M$ cells serve only as diagnostic anchors, not as part of the main fitted law.

\subsection{Compute Cost and Hardware Choice}
\label{sec:cost}

A sweep of hundreds of runs makes training cost part of the research design itself: hardware should be chosen by the cost of useful work, not by the hourly price alone. Before the main sweep, we measured the cost of an identical TinyStories run on different GPUs. The target metric is the cost of processing one million training tokens:
\begin{equation}
c_{1M} = \frac{\pi \cdot t}{3600 \cdot N_{\text{tok}}} \times 10^6, \label{eq:cost}
\end{equation}
where $\pi$ is the hourly GPU price, $t$ is the wall-clock run time, and $N_{\text{tok}}$ is the number of training tokens.

\begin{table}[h]
\centering
\caption{Cost and running time of an identical TinyStories run on an A10 and a T4 GPU. Prices are quoted in Russian rubles (RUB), reflecting the cloud provider used for these experiments.}
\label{tab:cost}
\begin{tabular}{lccccc}
\toprule
GPU & $\pi$, RUB/h & $t$, s & $N_{\text{tok}}$ & $c_{1M}$, RUB & Val.\ loss \\
\midrule
A10 & 36.55 & 221.97 & 40{,}960{,}000 & 0.055019 & 2.471797 \\
T4  & 21.75 & 1769.49 & 40{,}960{,}000 & 0.261002 & 2.470558 \\
\bottomrule
\end{tabular}
\end{table}

The A10 is roughly $7.97\times$ faster than the T4, at a $1.68\times$ difference in hourly price; per token processed, the A10 is roughly $4.74\times$ cheaper. Final validation loss is essentially identical, so the difference lies purely in cost and speed, not quality. This measurement determined how the main sweep was organized: heavier cells were run on the more performant GPU, while the T4 was reserved for control and infrastructure runs.

\section{Conclusion}
\label{sec:conclusion}

We tested Step Law in a regime the original paper never validated empirically, $N < 59M$. On pooled data from 935 runs, direct transfer of the original coefficients is rejected: Step Law systematically overestimates the optimal learning rate, by a median factor of about four. At the same time, the power-law form itself is preserved: $\eta^*$ is well described by a power law in $N$ and $D$, and $B^*$ by a power law in $D$ alone, with no significant contribution from $N$. The project's final answer is therefore H2: at small scale, Step Law needs recalibration, but not a different functional class.

The practical recommendation for small LLMs is to avoid using the Step Law formula verbatim. In our nanoGPT/TinyStories/BPE-2048 regime, it is more sensible to either use the recalibrated law $\hat\eta^* \approx 0.1\, N^{-0.51} D^{0.24}$, or, as a rough heuristic, to divide the learning rate predicted by Step Law by about four. For batch size, the authors' key structural conclusion holds: $B^*$ is primarily set by the data budget $D$, not by model scale $N$, although the slope with respect to $D$ is substantially steeper in our range.

The project's final artifact is a reproducible benchmark: code, configurations, run logs, tables of optima, and the final refit scripts, available at \url{https://github.com/kunikrubika05/step-law-small-scale}.

\section*{Reproducibility Statement}
All code, training configurations, per-run logs, and the scripts used for the final refit and figures are publicly available at \url{https://github.com/kunikrubika05/step-law-small-scale}. Table~\ref{tab:full-optima} in Appendix~\ref{app:optima} lists the robust optimum $(\hat\eta,\hat B)$, the number of completed runs, and the Step Law overshoot for every one of the 29 $(N,D)$ cells collected in this study.

\newpage

\bibliographystyle{plainnat}
\bibliography{references}

\newpage

\appendix
\section{Per-Cell Optima}
\label{app:optima}

Table~\ref{tab:full-optima} lists the robust-optimum estimates for all 29 unique $(N,D)$ cells collected across both iterations, including the two $N{=}5M$ anchor cells and the two high-$D/N$ diagnostic cells that are excluded from the main refit (marked with $\dagger$ and $\ddagger$, respectively). $\hat\eta$/$\hat B$ are the robust 2D-quadratic optima; $\eta_{SL}$ is the Step Law prediction $1.79\cdot N^{-0.713}\cdot D^{0.307}$; overshoot is $\eta_{SL}/\hat\eta$.

\begin{longtable}{lccccccc}
\caption{Robust-optimum estimates for all 29 unique $(N,D)$ cells.}
\label{tab:full-optima} \\
\toprule
Cell & Iter. & \#runs & $\hat\eta$ & $\hat B$ & $\hat L$ & $\eta_{SL}$ & Overshoot \\
\midrule
\endfirsthead
\toprule
Cell & Iter. & \#runs & $\hat\eta$ & $\hat B$ & $\hat L$ & $\eta_{SL}$ & Overshoot \\
\midrule
\endhead
N255K\_D8.39M   & it2 & 25 & 0.0085 & 1{,}024  & 2.851 & 0.0334 & $3.9\times$ \\
N255K\_D16.8M   & it2 & 30 & 0.0096 & 2{,}760  & 2.578 & 0.0413 & $4.3\times$ \\
N255K\_D33.6M   & it1 & 40 & 0.0105 & 3{,}697  & 2.361 & 0.0511 & $4.9\times$ \\
N255K\_D67.1M   & it2 & 25 & 0.0096 & 4{,}100  & 2.215 & 0.0632 & $6.6\times$ \\
N255K\_D134M    & it1 & 30 & 0.0140 & 8{,}190  & 2.144 & 0.0782 & $5.6\times$ \\
N255K\_D537M$^\dagger$    & it1 & 35 & 0.0318 & 53{,}912 & 2.027 & 0.1197 & $3.8\times$ \\
N519K\_D8.39M   & it2 & 25 & 0.0050 & 1{,}020  & 2.738 & 0.0201 & $4.0\times$ \\
N519K\_D16.8M   & it2 & 30 & 0.0088 & 2{,}050  & 2.395 & 0.0249 & $2.8\times$ \\
N519K\_D33.6M   & it1 & 30 & 0.0077 & 4{,}100  & 2.164 & 0.0308 & $4.0\times$ \\
N519K\_D67.1M   & it2 & 30 & 0.0094 & 7{,}883  & 2.041 & 0.0381 & $4.0\times$ \\
N519K\_D134M    & it1 & 30 & 0.0113 & 19{,}950 & 1.942 & 0.0471 & $4.2\times$ \\
N519K\_D268M    & it2 & 35 & 0.0141 & 31{,}395 & 1.876 & 0.0583 & $4.1\times$ \\
N519K\_D537M$^\dagger$    & it1 & 30 & 0.0229 & 58{,}324 & 1.819 & 0.0721 & $3.1\times$ \\
N997K\_D8.39M   & it2 & 25 & 0.0044 & 1{,}020  & 2.591 & 0.0126 & $2.9\times$ \\
N997K\_D16.8M   & it2 & 25 & 0.0065 & 2{,}475  & 2.277 & 0.0156 & $2.4\times$ \\
N997K\_D33.6M   & it1 & 30 & 0.0055 & 3{,}204  & 2.073 & 0.0193 & $3.5\times$ \\
N997K\_D67.1M   & it2 & 30 & 0.0045 & 6{,}922  & 1.937 & 0.0239 & $5.4\times$ \\
N997K\_D134M    & it1 & 35 & 0.0084 & 18{,}876 & 1.846 & 0.0296 & $3.5\times$ \\
N997K\_D268M    & it2 & 60 & 0.0059 & 21{,}755 & 1.745 & 0.0366 & $6.2\times$ \\
N997K\_D537M    & it1 & 35 & 0.0137 & 54{,}339 & 1.672 & 0.0453 & $3.3\times$ \\
N2.03M\_D8.39M  & it2 & 40 & 0.0027 & 1{,}020  & 2.457 & 0.0076 & $2.8\times$ \\
N2.03M\_D16.8M  & it2 & 40 & 0.0023 & 2{,}050  & 2.155 & 0.0094 & $4.1\times$ \\
N2.03M\_D33.6M  & it1 & 35 & 0.0041 & 2{,}050  & 1.913 & 0.0117 & $2.8\times$ \\
N2.03M\_D67.1M  & it2 & 40 & 0.0043 & 6{,}664  & 1.793 & 0.0144 & $3.4\times$ \\
N2.03M\_D134M   & it1 & 30 & 0.0044 & 8{,}190  & 1.700 & 0.0178 & $4.0\times$ \\
N2.03M\_D268M   & it2 & 25 & 0.0075 & 39{,}507 & 1.624 & 0.0220 & $2.9\times$ \\
N2.03M\_D537M   & it1 & 34 & 0.0088 & 46{,}584 & 1.544 & 0.0273 & $3.1\times$ \\
N5M\_D67.1M$^\ddagger$    & it2 & 25 & 0.0035 & 5{,}663  & 1.665 & 0.0076 & $2.1\times$ \\
N5M\_D134M$^\ddagger$     & it2 & 25 & 0.0031 & 8{,}921  & 1.557 & 0.0094 & $3.0\times$ \\
\bottomrule
\end{longtable}
\vspace{-0.5em}
{\small $^\dagger$ excluded from the main refit ($D/N>600$). \quad $^\ddagger$ $N{=}5M$ anchor cells, excluded from the main refit.}

\end{document}